**From task structures to world models: What do LLMs know?**

Ilker Yildirim[1,3,4,5] and L.A. Paul[2,4,5]
{ilker.yildirim, la.paul}@yale.edu
[1] Department of Psychology, Yale University
[2] Department of Philosophy, Yale University
[3] Department of Statistics & Data Science, Yale University
[4] Wu-Tsai Institute, Yale University
[5] Foundations of the Data Science Institute, Yale University

Sep 4 2023

**Abstract**

In what sense does a large language model have knowledge? The answer to this question extends beyond the capabilities of a particular AI system, and challenges our assumptions about the nature of knowledge and intelligence. We answer by granting LLMs "instrumental knowledge"; knowledge defined by a certain set of abilities. We then ask how such knowledge is related to the more ordinary, "worldly" knowledge exhibited by human agents, and explore this in terms of the degree to which instrumental knowledge can be said to incorporate the structured world models of cognitive science. We discuss ways LLMs could recover degrees of worldly knowledge, and suggest such recovery will be governed by an implicit, resource-rational tradeoff between world models and task demands.

**Introduction**

OpenAI's GPT-4[1], and similar large language models such as Meta's LLaMA[2], show impressive conversational capabilities. These systems can generate coherent, novel, and often surprisingly sophisticated responses to questions or prompts posed directly in natural language. Reflecting this perspective, often implicitly, artificial intelligence (AI) researchers that develop these systems[3], along with authors of articles in news media and scientific works, typically focus discussion on whether large language models have knowledge.

The goal of this opinion article is to explore this possibility. We suggest we are in a "Kuhnian moment" – a conceptual revolution in what we take knowledge to involve, with implications for how we think intelligence could arise. Accordingly, we ask: in what sense can GPT-4 (and similar models) be said to have knowledge? The answer to this question extends far beyond the capabilities of a particular AI chatbot, with implications for the fields of cognitive science, neuroscience, philosophy, and certainly AI.

We ground our answer using a core concept from cognitive science – world models[4]: causal abstractions of the entities and processes in the real world that preserve their structure, including objects with three-dimensional shapes and physical properties[5], scenes with spatial structure and navigable surfaces[6], and agents with beliefs and desires[7]. Human thought relies on these types of world models[8] to infer how physical processes work[5], and, importantly, to effectively reason[9], plan[10], and talk about the world[11].

In particular, for a wide range of ordinary contexts, a knowledgeable human agent draws on their world model, exploiting a structural match between their mental representations and the state of the world, to reliably generate contentful answers to prompts that are approximately truth-preserving and relevant. We will describe such world-model-based knowledge as "worldly" knowledge, and the content of the matched representations that support it as "worldly content". When a subject has worldly knowledge of a proposition, this is, at least in part, in virtue of the subject's using their world model to grasp the worldly content indicated in the proposition. For example, when the subject knows that *balancing a ball on a box is easier than balancing a box on a ball*, they use their world model of how objects move and react to external forces to grasp that balancing a ball on a box is easier than balancing a box on a ball. We take such knowledge to be the target of philosophical analyses of how an individual knows a proposition *p*, where to know that *p*, "an agent must not only have the mental state of believing that *p*, but various further independent conditions must also be met: *p* must be true, the agent's belief in *p* must be justified or well-founded, and so forth." [p. 281, Nagel 2013: "Knowledge as a Mental State", in Oxford Studies in Epistemology 4 (2013), 275-310] We also take such knowledge to be well-studied in scientific contexts[12–17], and to include much of our ordinary factual and relational knowledge about the world.

Recent developments in Artificial Intelligence (AI) seem to have created a different kind of knowledge. Such systems are based on deep neural networks pre-trained on Internet-scale data to autocomplete the next word (or token, more accurately) given preceding context, which are then further fine-tuned with reinforcement and supervised learning techniques for human-aligned and humanlike responses (e.g., Ouyang et al. 2022 [18]). When, using these tools, LLMs can generate sufficiently successful responses to an appropriately wide range of prompts, giving answers that are often approximately truth-preserving and relevant, we will describe such answers as demonstrating "instrumental knowledge."

This extends our question. We can grant that LLMs exhibit knowledge, for they can be said to have instrumental knowledge. But how is such instrumental knowledge related to ordinary human knowledge? As we will ask: to what degree, if any, can an LLM's instrumental knowledge incorporate worldly knowledge?

**Instrumental knowledge**

We can understand the instrumental knowledge of a system in terms of its ability to perform tasks posed for it across relevant domains. Indeed, a motivating perspective on large language models is the idea of unsupervised multitask learning [19]. Internet-scale natural language data can be seen as a large dataset of a multitude of tasks, posed in varying ways and forms, consistent with the messiness of how language is used naturally. For instance, the abbreviation "TL;DR" or a paragraph that starts with the phrase "In summary, …" might signal a summarization-like task; nearby or paired sentences or phrases spanning multiple languages might suggest the task of translation between those languages. Radford et al. (2019) [19] speculated that for a model to accurately predict the next word in a sequence, it may be critical for the model to spontaneously infer the task structure from the preceding context, and condition the next word predictions on that task structure. After the training of an LLM is over, inferring such task structure from natural language, and conditioning the activations within the model according to this structure, is a possible source of instrumental knowledge (Fig. 1A, B).

Could instrumental knowledge occur without (or with very little) worldly knowledge? A plausible example of such a scenario is machine language translation. Instead of focusing on building systems that translate through semantic analyzers or any other formal notion of meaning, most progress in machine translation relies on increasingly sophisticated statistical approaches.[20,21]

It is plausible that LLMs represent a new frontier in this progression of models, one in which the models infer the task structure of language translation, in terms of how words, phrases, and even paragraphs are emitted within and across pairs of languages, and use this structure to translate – without necessarily projecting a common knowledge of the physical world across different languages. Such a possibility is further suggested by "relational" theories of word meaning (e.g., such as a conceptual role semantics, where the meaning of a word or phrase is defined directly by its relation to other words or concepts, and only indirectly through reference and causal connections to the nonlinguistic world, with limited transmission of worldly content)[22].

Relatedly, some have asked whether LLM "knowledge" is merely an ability to follow linguistic rules and language patterns. In addressing this question, an important distinction with reference to the performance of LLMs is between 'form' vs. 'meaning'. Bender & Koller [23] rightfully caution that LLM's ability to generate coherent language should not be taken as evidence of natural language understanding – which in the present article, we operationalize such understanding in terms of having knowledge of the physical world.

Similarly, based on the separation observed in the human brain between language vs. non-language regions,[24,25] Mahowald et al. [26] argue that LLMs acquire formal linguistic competence, or knowledge of the rules[27] and statistical regularities [28] of a language, but not "functional linguistic competence", which includes knowledge of and reference to things and processes in the social and physical worlds[29]. In doing so, at least in part, Mahowald et al. draw on a distinction between pre-trained LLMs (on next-word prediction) vs. LLMs that are further fine-tuned with supervised and reinforcement learning objectives on human dialogue data (e.g., [18]).

Some fine-tuned LLMs, including GPT-4, when prompted with examples that seem to require worldly knowledge, including some of the examples Mahowald et al. considered, generate compelling answers. Such fine-tuning procedures only adapt certain output stages of the pre-trained models, otherwise keeping much of the pre-trained weights frozen[30,31]. Thus, explaining away aspects of the LLM performance that seem to go beyond formal competence as entirely a consequence of fine-tuning is difficult. Moreover, in certain cases, performance similar to that of a fine-tuned model can be obtained with pre-trained LLMs via so-called

"in-context learning", a form of learning where, in the absence of any parameter updates in the underlying model, simply providing example input-output pairs in the prompt leads LLMs to learn new tasks [32]. "Grounding" pre-trained LLMs (e.g., establishing a correspondence between a non-language domain and LLM embeddings) is an example of what can be accomplished with this kind of in-context learning [33].

This indicates to us that pre-trained LLMs may absorb bits and pieces of knowledge that give it abilities that go beyond what can be described as formal linguistic competence. For this reason, the instrumental knowledge that we ascribe to LLMs carves out a space that is distinct from and exceeds the rules and patterns of a language, and includes inferring a deeper task structure and conditioning next word prediction on that structure (Fig. 1A, 1B).

**The leap: From next word prediction to world models**

Instrumental knowledge, to count as "knowledge" in the ordinary sense, must somehow be about the world; it must include some degree of worldly content, in order to include some degree of worldly knowledge. But how could a purely text-related objective (i.e., next word/token prediction) lead to any degree of worldly knowledge (Fig. 1C)?

**A**

**Prompt:** Which one is easier: balancing a ball on a box or balancing a box on a ball?
**GPT-4:** Balancing a ball on a box is generally easier than balancing a box on a ball. (10/10)

**Prompt:** Which one is easier: balancing a sphere on a cube or balancing a cube on a sphere?
**GPT-4:** Balancing a cube on a sphere is generally easier than balancing a sphere on a cube. (8/10)

**B** **Instrumental knowledge:**
Inferring and using task structure

Pr(next word | input,
　　　　language rules and patterns,
　　　　`compare plausibility/frequency`)

**C** **Worldly knowledge:**
Inferring and using world models

Pr(next word | input,
　　　　language rules and patterns,
　　　　task structure,
　　　　`force-dynamic rules`)

Fig. 1. Exploring the nature of knowledge in large language models. (A) LLMs, such as OpenAI's GPT-4, show impressive capabilities on a broad range of tasks, and some failures, that collectively suggest questions to the extent of their worldly knowledge. (GPT-4 was queried 10 times on each of these prompts and the most frequent response – shortened for conciseness – is reported, with frequency displayed in green if that response is consistent with the authors' intuitions and red otherwise.) (B) We suggest that pre-trained LLMs acquire instrumental knowledge that goes beyond formal linguistics competence (i.e., the set of rules and statistical patterns that makes up language). We suggest that this falls from the objective of next word prediction – accurate auto-completion leads to inferring and using task structure during the processing of input context – and is perhaps further strengthened and organized during the fine-tuning of LLMs. This pair of prompts illustrate an example, where the model might infer that the task is to compare the plausibility of different phrases in the prompt, with respect to the (often highly complex) co-occurrence patterns of these phrases. Such instrumental knowledge might work for comparisons of phrases whose co-occurrence patterns are approximated well by the model, but become increasingly inaccurate for less frequent phrases. (C) We explore how much of this instrumental knowledge might draw on an underlying account of the entities and processes in the physical world –

which we term worldly knowledge. In this example, this would include the force-dynamic relations between entities. We examine this perspective via the core cognitive concept of structured world models, a review of recent relevant work, and the notion of bounded rationality.

One interesting possibility draws on the ability to recover worldly content using compression.[34,35] As observed by Chiang [36], next word prediction in LLMs reflects compression of the vast amounts of text data crawled on the Internet into the many billions of weights of a deep neural network, which in proportion remain too small to memorize the training data. Indeed, compression and prediction are closely related objectives.[34,35]

A lower-dimensional state-space factorizing the relevant dimensions of variation of a given domain and the dependence of these dimensions on each other can simultaneously enable compression and prediction. Structured world models are examples of such state-spaces, where a small set of variables captures causal abstractions of the structure of their counterpart physical processes in the real world (Fig. 2A). The caveat, however, is that there are many ways a dataset can be compressed, leading to various approximations of the statistical regularities in the training text.

However, we speculate that compression could recover the data generating process underlying the training data. There are certainly multiple data generating processes underlying natural language data, including the rules and statistical regularities in a language, how tasks are posed and addressed, and especially relevant to the present context, worldly content involving the entities, physical processes, and situations projected into text by the humans perceiving and participating in these situations and talking about them.

We acknowledge that, at present, whether LLMs actually do or even could recover causal abstractions of the world, as in world models, involves a leap of faith. However, as it is often this possibility that motivates attributions of general intelligence to LLMs (e.g., Bubeck et al. [37]), we turn to available literature exploring this possibility.

**Measured recovery of world models in LLMs under domain-specific settings**

We think that structured world models – i.e., causal abstractions of a represented system with a structure preserving (homomorphic) representing system[38] (Fig. 2A) – provide a concrete framework to reason about whether and how LLMs can recover degrees of worldly content that could lead to worldly knowledge.

A small number of recent studies have explored whether a language model trained to predict next-token spontaneously approximates the underlying data generating process – i.e., the world model.[39,40] To do so, these studies moved away from the natural language setting to specific domains in which input sequences are still text-based and can be tokenized just like language, but each token reflects the unfolding of an underlying domain-specific world model.

One example of the recovery of worldly knowledge is presented by Li et al. [39] (Fig. 2B). Li et al. trained a GPT-4-like (but on a much smaller scale) language model to predict legal moves in the game of Othello – a two-player board game in which players gain more discs by outflanking the discs of the other player. The world model in this game consists of the state of the board (for each cell in an 8x8 grid, whether it is black, white, or empty) and the set of rules by which each player changes this state. Li et al. randomly simulated this world model creating a dataset of board state traces via stochastic, non-strategic decisions for each player. When they trained a language model, called Othello-GPT, on this dataset, they found that not only this model could reliably generate legal moves given the prior set of moves, but also the entire board state could be accurately decoded from the intermediate-layer activations in the model, with a linear decoder (as established by a follow-up study [41]). Crucially, the authors also showed that intervening on the board state via these decoders causally and appropriately impacted the model's legal move predictions.

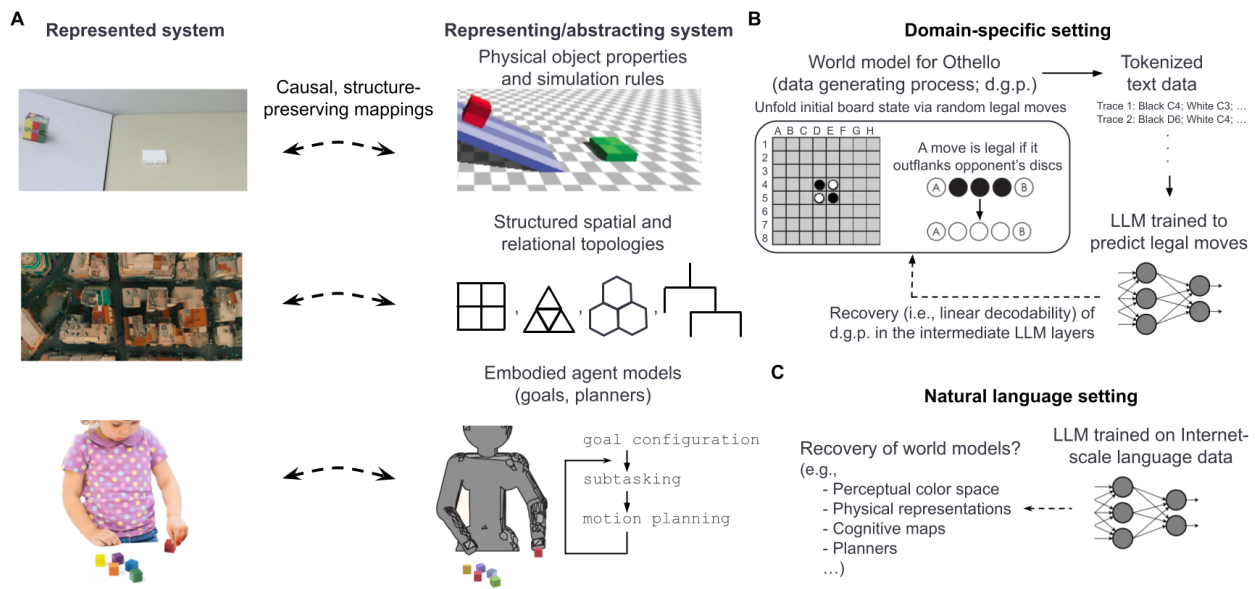

Fig. 2. Using the core concept of world models from cognitive science to explore the extent of worldly knowledge in LLMs. (A) World models in the example domains of intuitive physics, cognitive maps, and embodied planning. World models are homomorphic mappings that represent, in an abstract format, objects with physical properties, places with navigable spatial relations, and agents with beliefs and goals. (B) Compression and prediction are like the two sides of a coin, and it is possible compression can recover the data generating process. In a domain-specific setting (the board game of Othello) with tokenized traces of randomly generated game states, Li et al. (2022) trained an LLM on next-token prediction. Surprisingly, the intermediate layers of this LLM yielded a linearly decodable full board state. (C) As we discuss in the text, generalization of this result to actual natural language trained LLMs is so far limited.

Furthermore, Jin & Rinard (2023) [40], using a toy domain, took a similar approach to suggest that a language model trained for next-token prediction for program synthesis also recovers something about the deeper semantics of this domain-specific programming language. Finally, a similar conclusion comes from the domain of computational biology: a recent LLM trained by Meta researchers [42] to predict masked entries in protein amino acid sequences rendered the coarse three-dimensional structure (i.e., contact relations between amino acids) of the actual folded protein linearly decodable.

The work of Li et al. [39] establishes the speculative possibility mentioned above: that next-token prediction with a language model can recover the underlying data generating process. It suggests that the "leap" mentioned in the previous section is in fact a realistic outcome. It is exciting and of pressing importance for future research to systematically explore this possibility across the dimensions of training objective (e.g., next-token prediction, masked token prediction), network architectures (e.g., transformer based language models[43], RNN-based sequence models[44,45] – the kinds of neural network architectures that underlie the modern LLMs and traditional neural language models, respectively), and the complexity of world models.

That said, we raise three caveats. First, these studies depend on dense sampling of relatively small domains (more than 1 million training examples for Othello); dataset requirements can quickly grow for non-toy domains, and whether the largest Internet-scale language datasets can satisfy these conditions is not known. Second, these studies consider settings where the basic building blocks of the world models can be enumerated and assigned unique tokens – e.g., the locations in the 8x8 grid and the possible states for each cell in the Othello environment. We suspect this will not be applicable for many relevant world models and for how they are projected into text by human language users. Finally, it is possible that the recovered world model in a language model, based on the objective of next-token prediction, might be computationally costly.

This motivates the exploration of the recovery of domain-specific world models in LLMs trained with natural language data (Fig. 2C),[33,46,47] drawing on recent work in the context of perceptual color space.[33,46]

Abdou et al. [46] shows that pre-trained language models can recover aspects of the relational structure of the perceptual color space. Using representational similarity analysis, they report statistically significant correlations between the similarity structure of color pairs with respect to language model embeddings vs. euclidean distances between the same pairs of colors under a well-established perceptual color space. Patel & Pavlick (2022) [33] provides further evidence (using larger language models, including GPT-3[32]) but with a different approach: They report that LLMs provided with a small number of in-context examples from a single hue (e.g., red) generalize, more accurately than chance, to the rest of the color space, indicating that some aspects of the relational structure of the perceptual color space is readily available in these LLMs.

In a way, these results suggest the possibility of recovering worldly knowledge in LLMs, despite their purely text-based training. But this needs to be qualified. The quantitative nature of the correspondence between the physical spaces and LLM internals is often underwhelming in these studies (e.g., a correlation value of roughly 0.2 in the Abdou et al. study). We anticipate that this correspondence will increase under better trained, larger models, but nevertheless, the fact that this relationship is weak in a domain like colors is telling: structure-wise, color space is a simple topology (distances in 3D space) and presumably there is much text in the training corpora that talks about color. The complexity of typical world models projected to text by human language users (e.g., spatial structures, intuitive physics) is often far more complex. It is plausible that in most realistic settings, LLMs will primarily acquire instrumental knowledge that reflects only limited or partial worldly knowledge, much like the case of machine translation.

**A resource rational view of instrumental knowledge**

As we noted above, worldly knowledge can be costly – some of these mappings illustrated in Fig. 2A (for example, a physics engine, embodied planning and manipulation) may rival LLMs in the complexity of the computations they involve, as they capture, in algorithmically efficient forms, complex causal processes from the physical world.

Fortunately, for most tasks, partial or coarser worldly knowledge is good enough. To see an example where coarser worldly knowledge would be sufficient, consider the domain of intuitive physics – our ability to predict, often at a glance, how objects will move and react to external forces.[48,49] An influential cognitive computational framework suggests that such predictions arise from structured world models in the mind[9]: runnable mental models of physical objects implemented using appropriate homomorphic representations, e.g., simulations in probabilistic or approximate forms of physics engines from computer graphics (Fig. 2A, top).

The mental precision of such a simulation can vary with task requirements[50,51]: A coarse-grained, qualitative simulation might suffice to predict whether a liquid will flow toward right or left, while a finer-grained simulation may be necessary to work out details of its trajectory [52,53]. Representation granularity provides a direct measure over the extent of worldly knowledge, and the task demands on world models can provide a useful resource-rational[54–56] perspective on the kind of instrumental knowledge demonstrated in LLMs.

Thus, we suggest the degree to which worldly knowledge is part of instrumental knowledge in LLMs may be determined in a resource-rational way, stemming from a tradeoff between a need for (costly) worldly knowledge and the need to perform the tasks that occur frequently during training. Instrumental knowledge can be sufficient for accurate prediction in scenarios where a coarse-grained mental simulation suffices, but such prediction may become increasingly inaccurate in scenarios that require finer-grained simulation in the mind (also see Jara-Ettinger & Rubio-Fernandez[57]). An important caveat here is that in LLMs, this tradeoff occurs implicitly, with its black-box and fundamentally difficult to control composition.

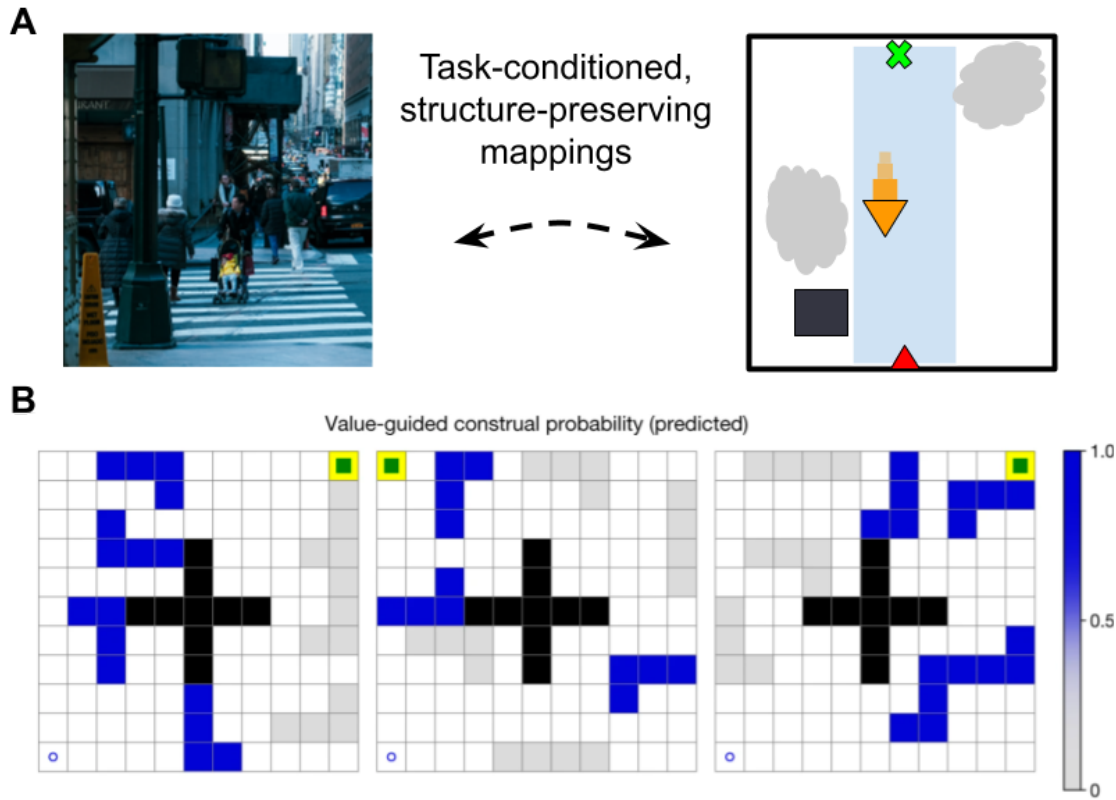

Fig. 3. Resource-rational world models that reflect goals of an agent. (A) A task-driven homomorphic mapping would invest less resources on the aspects of worldly knowledge not benefiting the task of the agent, but otherwise would still be a causal abstraction. (B) Ho et al. [58] introduced a framework that allows constructions of simplified task representations by balancing the complexity of task representations with the expected value of plans. For instance, given a maze navigation task, including a starting location (blue circle), a target location (highlighted cell), and the walls (rest of the non-white cells), the model constructs a simplified representations in which only a subset of the walls are likely to be included, while maintaining the expected accuracy and path length of the plan. From Ho et al. (2022).

Future AI systems, as well as future empirical explorations of LLMs, should aspire to directly address this tradeoff by integrating task demands during the formation of world models (Fig. 3A). Indeed, humans make this tradeoff between knowing about the world and their goals all the time, leading to structured, but strikingly sparse percepts.[59–61] Such task-driven mental representations carve just enough of the complexity of the physical world to enable the (momentary) goals of an agent[61]. It is only recently that the field of cognitive science has started exploring computational theories that integrate structured world models with task demand[58]. For example, Ho et al. [58] provided a computational-level explanation of simplified representations relative to planning objectives, using the domain of 2D maze navigation (Fig. 3B). This new landscape of resource-rational world models could in turn help refine what we take knowledge to consist in.

**World models as programmable, mid- or high-level interfaces for safety and alignment**

Beyond their role in intelligence, why else should we embrace the use of world models in an AI system? That is, if a system has instrumental knowledge allowing it to perform a diverse range of tasks accurately, what else should we want?

We want such systems to be safe. That is, we want these AI systems to be deployed in a way that is both truthful and aligned with "human values."[62] Existing deep neural networks, including LLMs with their transformer based neural network architectures, are black-box systems without an explicit, high-level interface for directly programming their behavior. The lack of transparency in how LLMs work, combined with their tendency to produce "hallucinations" – the frequently observed fabrication of situations, events, and persons by these systems in response to reasonable prompts – has raised concerns in the industry and in larger societal contexts.[63,64] The possibility of unexpected value change in these systems (for example, the higher order value change detailed in Paul (2014) [65]) raises questions about how sure we can be that evolving AI values will remain aligned with human values. [62]

In contrast, world models offer a path to safer, truthful, and better aligned AI systems. Because world models, and more generally domain-specific high-level programming languages, formalize worldly knowledge in structure-preserving, interpretable representations, they can readily enable truthfulness, supporting an engineer or a user who wishes to impart their "values" and safety measures as explicit constraints over the system. These features can be exploited via hybrid pipelines of LLMs and world models, [66–68] neuro symbolic architectures [69], and by creating natively programmable neural network architectures. [70]

**Conclusion**

The impressive performance of LLMs on a surprisingly broad range of tasks challenges how we think about the nature of knowledge, and by extension, how intelligence could arise in artificial or machine systems. To frame this challenge, we asked: in what sense can LLMs, trained purely on text, primarily to predict the next word, be said to have knowledge? We answered the challenge by granting "instrumental knowledge" to LLMs: knowledge defined by a certain set of abilities, including particular linguistic and task-based abilities, and suggested this may be a new kind of knowledge. We then asked how such knowledge is related to the more ordinary, "worldly" knowledge exhibited by people, and explored this in terms of the degree to which instrumental knowledge can be said to incorporate world models.

We close by noting that instrumental knowledge in LLMs could involve an implicit resource-rational tradeoff between the otherwise costly use of structured world models and the need to perform commonly occurring tasks. Future work should construct goal-conditioned world models, structured abstractions that nevertheless can integrate the agent's planning objectives. Beyond their implication for intelligence, world models, incorporated more explicitly in AI systems, will facilitate a more direct path safe and aligned deployment, by exposing an interpretable mid- or high-level interface for control and intervention (see Outstanding Questions).

**Acknowledgements**

We thank Yutaro Yamada, Mario Belledonne, and Jonathan Schaffer for helpful discussions. We thank members of the Cognitive & Neural Computation Lab at Yale for their feedback on earlier versions of this manuscript. We thank Aalap Shah for his help making display items. I.Y. was supported by AFOSR grant # FA9550-22-1-0041.

**Glossary**

Instrumental knowledge: knowledge of true propositions had in virtue of a system's set of abilities. For an LLM, this includes its ability to perform tasks posed for it across a wide range of relevant domains, including the ability to provide coherent and often surprisingly sophisticated responses to natural language prompts. We suggest that such performance rests on an LLM's ability to spontaneously infer task structure from natural language input, and to condition the activations within the model according to this structure for next word prediction. S demonstrates instrumental knowledge of P: "balancing a ball on a box is easier than balancing a box on a ball", when it can spontaneously infer P in response to a natural language prompt (in addition to having the more generally defined set of abilities detailed above).

Worldly content: content constituted by entities and processes in the real world, including objects with three-dimensional shapes and physical properties, scenes with spatial structure and navigable surfaces, and agents.

Worldly knowledge: S has worldly knowledge *that P* only if P has worldly content and S uses their world model to grasp this content.

Ordinary knowledge: a subject S knows proposition P (in the ordinary way) in virtue of standing in the knowledge relation to the (true) proposition P.

Homomorphic mapping: A mapping between two systems is homomorphic if the structure of how the symbols within each system relate to each other is preserved. A basic example of homomorphism is linearly mappable systems.

World models: Even though world models are talked about in different ways in AI literature, we refer to a specific concept grounded in representation theory: homomorphic (i.e., structure preserving) representations of real-world processes. These representations are causal in the sense that they capture, at an abstract level, their counterpart real-world processes, in algorithmically efficient forms, such as simulating optics with computer graphics and object motion with physics engines. World models can be embedded within probabilistic models to specify inference, prediction, and reasoning queries, to formalize relevant processes in perception and cognition.

Pre-trained LLMs: Deep neural networks trained to autocomplete text on Internet-scale natural language data. A typical LLM will have hundreds of millions to hundreds of billions tunable parameters. The currently best performing LLMs share a common network architecture known as the "transformer" architecture. Transformer architecture is unique in the way it allows the models to process a long context of words at once, and attend over that context in ways that are highly flexible.

Fine-tuned LLM: Pre-trained LLMs that are further trained, typically on much less data than used during pre-training, to follow instructions and generate aligned and humanlike responses, often using reinforcement and supervised learning techniques (although other methods are possible).

1. OpenAI. GPT-4 Technical Report. *arXiv [cs.CL]* Preprint at http://arxiv.org/abs/2303.08774 (2023).
2. Touvron, H. *et al.* LLaMA: Open and Efficient Foundation Language Models. *arXiv [cs.CL]* (2023).
3. Eye on AI. The Mastermind Behind GPT-4 and the Future of AI | Ilya Sutskever. https://www.youtube.com/watch?v=SjhIlw3Iffs (2023).
4. Lake, B. M., Ullman, T. D., Tenenbaum, J. B. & Gershman, S. J. Building machines that learn and think like people. *Behav. Brain Sci.* **40**, e253 (2017).
5. Yildirim, I., Siegel, M. & Tenenbaum, J. B. Physical Object Representations. in *The Cognitive Neurosciences, 6th edition* (ed. Poeppel, G. M.) 399 (MIT Press, 2020).
6. Epstein, R. A., Patai, E. Z., Julian, J. B. & Spiers, H. J. The cognitive map in humans: spatial navigation and beyond. *Nat. Neurosci.* **20**, 1504–1513 (2017).
7. Jara-Ettinger, J., Gweon, H., Schulz, L. E. & Tenenbaum, J. B. The Naïve Utility Calculus: Computational Principles Underlying Commonsense Psychology. *Trends Cogn. Sci.* **20**, 589–604 (2016).
8. Spelke, E. S. Core knowledge. *Am. Psychol.* **55**, 1233–1243 (2000).
9. Battaglia, P. W., Hamrick, J. B. & Tenenbaum, J. B. Simulation as an engine of physical scene understanding. *Proceedings of the National Academy of Sciences* **110**, 18327–18332 (2013).
10. Baker, C. L., Jara-Ettinger, J., Saxe, R. & Tenenbaum, J. B. Rational quantitative attribution of beliefs, desires and percepts in human mentalizing. *Nat. Hum. Behav.* **1**, 1–10 (2017).
11. Jones, C. R. & Bergen, B. The Role of Physical Inference in Pronoun Resolution. *Proceedings of the Annual Meeting of the Cognitive Science Society* **43**, (2021).

12. Dretske, F. Knowledge and the Flow of Information, Cambridge, MA: Bradford. Preprint at (1981).

13. Chisholm, R. M. *Theory of Knowledge*. (Prentice-Hall, 1977).

14. Goldman, A. I. *Epistemology and Cognition*. (Harvard University Press, 1986).

15. Kornblith, H. *Knowledge and Its Place in Nature*. (Clarendon Press, 2002).

16. Nagel, J. *Knowledge: A Very Short Introduction*. (OUP Oxford, 2014).

17. Williamson, T. *Knowledge and Its Limits*. (Oxford University Press, 2002).

18. Ouyang, L. *et al.* Training language models to follow instructions with human feedback. *Adv. Neural Inf. Process. Syst.* **35**, 27730–27744 (2022).

19. Radford, A. *et al.* Language models are unsupervised multitask learners. *OpenAI blog* **1**, 9 (2019).

20. Och, F. J., Gildea, D., Khudanpur, S. & Sarkar, A. A smorgasbord of features for statistical machine translation. *Proceedings of the* (2004).

21. Stahlberg, F. Neural Machine Translation: A Review. *jair* **69**, 343–418 (2020).

22. Piantasodi, S. T. & Hill, F. Meaning without reference in large language models. *arXiv preprint arXiv:2208.02957* (2022).

23. Bender, E. M. & Koller, A. Climbing towards NLU: On Meaning, Form, and Understanding in the Age of Data. in *Proceedings of the 58th Annual Meeting of the Association for Computational Linguistics* 5185–5198 (Association for Computational Linguistics, 2020).

24. Fedorenko, E., Behr, M. K. & Kanwisher, N. Functional specificity for high-level linguistic processing in the human brain. *Proc. Natl. Acad. Sci. U. S. A.* **108**, 16428–16433 (2011).

25. Fedorenko, E. & Thompson-Schill, S. L. Reworking the language network. *Trends Cogn. Sci.* **18**, 120–126 (2014).

26. Mahowald, K. *et al.* Dissociating language and thought in large language models: a cognitive perspective. *arXiv [cs.CL]* (2023).

27. Chomsky, N. *Aspects of the Theory of Syntax, 50th Anniversary Edition*. (MIT Press, 2014).

28. Bybee, J. & Hopper, P. Introduction to frequency and the emergence of linguistic structure. *torrossa.com*.

29. Clark, H. H. *Using Language*. (Cambridge University Press, 1996).

30. Hu, E. J. *et al.* LoRA: Low-Rank Adaptation of Large Language Models. *arXiv [cs.CL]* (2021).

31. Tsimpoukelli, M. *et al.* Multimodal few-shot learning with frozen language models. *Adv. Neural Inf. Process. Syst.* **34**, 200–212 (2021).

32. Brown, T. *et al.* Language models are few-shot learners. *Adv. Neural Inf. Process. Syst.* **33**, 1877–1901 (2020).

33. Patel, R. & Pavlick, E. Mapping Language Models to Grounded Conceptual Spaces. (2022).

34. Grunwald, P. A tutorial introduction to the minimum description length principle. *arXiv [math.ST]* (2004).

35. Ratsaby, J. Prediction by Compression. *arXiv [cs.IT]* (2010).

36. Chiang, T. ChatGPT is a Blurry JPEG of the Web. *New Yorker* (2023).

37. Bubeck, S. *et al.* Sparks of Artificial General Intelligence: Early experiments with GPT-4. *arXiv [cs.CL]* (2023).

38. Gallistel, C. R. & King, A. P. *Memory and the Computational Brain: Why Cognitive Science will Transform Neuroscience*. (John Wiley & Sons, 2011).

39. Li, K. *et al.* Emergent World Representations: Exploring a Sequence Model Trained on a Synthetic Task. *arXiv [cs.LG]* (2022).

40. Jin, C. & Rinard, M. Evidence of Meaning in Language Models Trained on Programs. *arXiv [cs.LG]* (2023).

41. Nanda, N. Actually, Othello-GPT Has A Linear Emergent World Model. *neelnanda. io* Preprint at (2023).

42. Lin, Z. *et al.* Evolutionary-scale prediction of atomic-level protein structure with a language model. *Science* **379**, 1123–1130 (2023).

43. Vaswani, A. *et al.* Attention is all you need. *Adv. Neural Inf. Process. Syst.* **30**, (2017).

44. Elman, J. L. Finding structure in time. *Cogn. Sci.* **14**, 179–211 (1990).

45. Hochreiter, S. & Schmidhuber, J. Long short-term memory. *Neural Comput.* **9**, 1735–1780 (1997).

46. Abdou, M. *et al.* Can Language Models Encode Perceptual Structure Without Grounding? A Case Study in Color. *arXiv [cs.CV]* (2021).

47. Søgaard, A. Grounding the Vector Space of an Octopus: Word Meaning from Raw Text. *Minds Mach.* **33**, 33–54 (2023).

48. Kubricht, J. R., Holyoak, K. J. & Lu, H. Intuitive Physics: Current Research and Controversies. *Trends Cogn. Sci.* **21**, 749–759 (2017).

49. McCloskey, M. Intuitive physics. *Sci. Am.* (1983).

50. Clark, A. Radical predictive processing. *South. J. Philos.* **53**, 3–27 (2015).

51. Drugowitsch, J., Wyart, V., Devauchelle, A.-D. & Koechlin, E. Computational precision of mental inference as critical source of human choice suboptimality. *Neuron* **92**, 1398–1411 (2016).

52. Bates, C. J., Yildirim, I., Tenenbaum, J. B. & Battaglia, P. Modeling human intuitions about liquid flow with particle-based simulation. *PLoS Comput. Biol.* **15**, e1007210 (2019).

53. Zhang, Y., Belledonne, M., Yates, T. & Yildirim, I. Where does the flow go? Humans automatically predict liquid pathing with coarse-grained simulation. *Proceedings of the Annual Meeting of the Cognitive Science Society* **45**, (2022).

54. Gigerenzer, G. & Selten, R. *Bounded Rationality: The Adaptive Toolbox*. (MIT Press, 2002).

55. Lieder, F. & Griffiths, T. L. Resource-rational analysis: Understanding human cognition as the optimal use of limited computational resources. *Behav. Brain Sci.* **43**, e1 (2019).

56. Gershman, S. J., Horvitz, E. J. & Tenenbaum, J. B. Computational rationality: A converging paradigm for intelligence in brains, minds, and machines. *Science* **349**, 273–278 (2015).

57. Jara-Ettinger, J. & Rubio-Fernandez, P. Quantitative mental state attributions in language

understanding. *Sci Adv* **7**, eabj0970 (2021).

58. Ho, M. K. *et al.* People construct simplified mental representations to plan. *Nature* **606**, 129–136 (2022).

59. Neisser, U. The control of information pickup in selective looking. in *Perception and its development* 201–219 (Psychology Press, 1979).

60. Rensink, R. A., O'Regan, J. K. & Clark, J. J. To See or not to See: The Need for Attention to Perceive Changes in Scenes. *Psychol. Sci.* **8**, 368–373 (1997).

61. Niv, Y. Learning task-state representations. *Nat. Neurosci.* **22**, 1544–1553 (2019).

62. Bowman, S. R. Eight Things to Know about Large Language Models. *arXiv [cs.CL]* (2023).

63. Critch, A. & Russell, S. TASRA: a Taxonomy and Analysis of Societal-Scale Risks from AI. *arXiv [cs.AI]* (2023).

64. Russell, S. Provably beneficial artificial intelligence. *27th International conference on intelligent user* (2022).

65. Paul, L. A. *Transformative Experience*. (Oxford University Press, 2014).

66. Wong, L. *et al.* From Word Models to World Models: Translating from Natural Language to the Probabilistic Language of Thought. *arXiv [cs.CL]* (2023).

67. Wolfram, S. Chatgpt gets its 'wolfram superpowers'. *https://writings.stephenwolfram.com/2023/03/chatgpt-gets-its-wolfram-superpowers/ de https://writings. stephenwolfram. com*.

68. Ellis, K. Modeling Human-like Concept Learning with Bayesian Inference over Natural Language. *arXiv [cs.CL]* (2023).

69. Lu, X., Welleck, S., West, P., Jiang, L. & Kasai, J. Neurologic a* esque decoding: Constrained text generation with lookahead heuristics. *arXiv preprint arXiv* (2021).

70. Kim, J. Z. & Bassett, D. S. A neural machine code and programming framework for the reservoir computer. *Nature Machine Intelligence* **5**, 622–630 (2023).

**Outstanding questions**

How can we formalize instrumental knowledge? One possibility is task-conditioned world models, which needs further development. Future work should also explore the relationship of instrumental knowledge to amortized inference or data-driven proposals in Bayesian inference and resource-rational solutions to intractable problems arising from working with expressive world models.

What is the impact of fine-tuning on an LLM's instrumental knowledge and recovery of worldly knowledge?

How can we incorporate structured world models in AI, such as LLMs and beyond, for safer and better aligned systems?

Empirical work should use world models and state-of-the-art computational theories in cognitive science to create domain-specific (e.g., intuitive physics, cognitive maps) benchmarks for assessing and monitoring the extent of worldly knowledge in LLMs. These benchmarks should explore dimensions such as world model granularity and the training data distribution.

LLMs are often referred to as "foundation models", in the sense of their adaptability to new tasks with little additional data. To what extent can the distinction of instrumental knowledge vs. worldly knowledge help toward building and exploring new foundation models in language and other fields, such as computer vision and reinforcement learning?

We assume that LLMs internalize formal linguistic competence. Is this fair? If so, what is the format of such knowledge, specifically in relation to linguistic theories?

How does our discussion relate to existing treatments of LLMs in related contexts, including conceptual role semantics and embodiment-based arguments?

We avoided using "understanding" in our discussion of knowledge in LLMs, as the term has multiple interpretations and is formally less developed. Does our discussion suggest an opportunity for analyzing "understanding" in terms of worldly knowledge?

**Highlights**

OpenAI's GPT-4, and similar large language models (LLMs), show impressive conversational capabilities.This article asks: In what sense does an LLM have knowledge? The answer to this question extends beyond the capabilities of a particular AI chatbot, and challenges our assumptions about the nature of knowledge and intelligence.

We answer by granting LLMs "instrumental knowledge" – knowledge defined by a certain set of abilities.

How is such knowledge related to the more ordinary, "worldly" knowledge exhibited by humans? To address, we turn to a core concept in cognitive science, world models, and explore the degree to which instrumental knowledge might incorporate such structured representations.

We discuss how LLMs could recover degrees of worldly knowledge, and suggest such recovery will be governed by an implicit, resource-rational tradeoff between world models and task demands.